\documentclass[journal]{IEEEtran}

\usepackage{hyperref}
\usepackage{array}
\newcolumntype{P}[1]{>{\centering\arraybackslash}p{#1}}

\ifCLASSINFOpdf
\else
   \usepackage[dvips]{graphicx}
\fi
\usepackage{url}

\usepackage{graphicx}
\usepackage{color}
\usepackage{amsmath}
\usepackage[flushleft]{threeparttable}

\usepackage{algorithm}
\usepackage{algorithmic}[1]

\begin{document}

\title{A Computationally Efficient Method for Defending Adversarial Deep Learning Attacks}

\author{Rajeev Sahay, \IEEEmembership{Student Member, IEEE}, Rehana Mahfuz, \IEEEmembership{Student Member, IEEE}, \\and Aly El Gamal, \IEEEmembership{Member, IEEE}
\thanks{R. Sahay, R. Mahfuz, and A. El Gamal are with the Department of Electrical and Computer Engineering, Purdue University, West Lafayette, IN, USA. Email: \{sahayr, rmahfuz, elgamala\}@purdue.edu.}}

\markboth{IEEE Signal Processing Letters}
{Shell \MakeLowercase{\textit{et al.}}: Bare Demo of IEEEtran.cls for IEEE Journals}
\maketitle

\begin{abstract}
The reliance on deep learning algorithms has
grown significantly in recent years. Yet, these
models are highly vulnerable to adversarial attacks,
which introduce visually imperceptible perturbations
into testing data to induce misclassifications. The literature has proposed several methods to combat such adversarial attacks, but each method either fails at high perturbation values, requires excessive computing power, or both. This letter proposes a computationally efficient method for defending the Fast Gradient Sign (FGS) adversarial attack by simultaneously denoising and compressing data. Specifically, our proposed defense relies on training a fully connected multi-layer Denoising Autoencoder (DAE) and using its encoder as a defense against the adversarial attack. Our results show that using this dimensionality reduction scheme is not only highly effective in mitigating the effect of the FGS attack in multiple threat models, but it also provides a 2.43x speedup in comparison to defense strategies providing similar robustness against the same attack. 
\end{abstract}


\IEEEpeerreviewmaketitle

\section{Introduction}

\IEEEPARstart{T}{he} recent emergence of adversarial machine learning attacks has exposed the vulnerability of artificial neural networks \cite{intregprops}. These attacks consist of corrupted data samples, termed \emph{adversarial examples}, which are constructed by adding visually imperceptible errors into testing data to induce high confidence misclassification on well-trained machine learning classifiers \cite{eattacks}. The Fast Gradient Sign (FGS) attack is one particularly fast, simple, and highly effective adversarial attack algorithm, which relies on adding, according to the sign of the cost function's gradient, an $l_\infty$-bounded error to each data sample resulting in high-confidence misclassification \cite{fgs}. As a result of this vulnerability, several defenses have been proposed for a variety of threat models to combat the FGS attack. Yet, prior defenses are either computationally costly, lack robustness in resisting misclassification at high perturbation magnitudes, or both. This letter proposes a defense against the FGS attack that mitigates its effects on deep learning classifiers in a wide perturbation range while retaining computational efficiency.


Although several types of adversarial attacks exist (see e.g., \cite{pgd}, \cite{deepfool}, and \cite{cw}), this letter's specific focus on defending the FGS attack is twofold. First, the currently proposed state-of-the-art defenses for combating this attack are computationally inefficient whereas less costly algorithms lack accuracy in classification. In the scope of this letter, we aim to fill this gap by devising a defense that is both computationally efficient and robust in resisting misclassification. Second, more potent adversarial attacks such as Projected Gradient Decent \cite{pgd}, DeepFool \cite{deepfool}, and the Carlini \& Wagner Attack \cite{cw}, are iterative and require a longer implementation time than the FGS attack. The consequential latency caused by these attacks is more likely to result in tamper detection \cite{ttm}, in real-time, and cause the corrupted sample to be rejected for classification. The FGS attack, however, is fast to implement and more likely to bypass latency measurements making it more desirable by an attacker. 

We examine the effectiveness of our proposed defense in two defense-blind attack environments: the semi-white box threat model, in which the attacker has full knowledge of the target classifier's architecture and parameters, and the black box threat model in which the attacker is aware of the classification task performed by the deep neural network but unaware of the target model's architecture and parameters. As will be clear in the sequel, black box environments are not necessarily less vulnerable than semi-white box environments due to the transferability property of the FGS attack across disparate classifiers \cite{transfer}. Therefore, the effectiveness of our proposed defense is tested in both scenarios. 

It is important to note that prior work \cite{ograds}-\cite{towardsdnn} has shown differentiable pre-processing functions to be vulnerable to further adversarial attacks (i.e., combining the defense with the target model and using its resulting gradient to generate an attack) in complete white box scenarios where the adversary has full knowledge of the defense. Although such methods can be circumvented in the white box scenario, the defense blind environment is equally important to consider as rapid software updates are capable of instantly changing the portion of the computational framework responsible for defending the attack. Therefore, unlike static Compact Disk software packages, present day downloadable software is highly susceptible to dynamic changes \cite{dynsoftupdating}, which can instantly alter an adversary's knowledge about a system. 

{\bf Related Work:} The majority of algorithms proposed for defending the FGS attack either rely on filtering, which consists of subtracting the potentially added adversarial noise, or dimensionality reduction, which consists of data compression. Examples of filtering include using the output of a Denoising AutoEncoder (DAE) (see e.g., \cite{towardsdnn} and \cite{dunet}) and manifold learning \cite{magnet}, which aims to detect corrupted samples by learning the distribution of the original uncorrupted training data and eliminating outliers. Although filtering using the output of a DAE is effective in mitigating the FGS attack, it requires the reconstruction of the input sample's compressed representation which is unnecessary, as we show in this work, not only because similar robustness can be achieved against the same attack using the compressed representation, but also because the compressed representation could reduce the training time of the defense. Furthermore, manifold learning has been shown to be vulnerable to the FGS attack, as adversarial examples may not appear as statistical outliers on the manifold of the training data \cite{magnetnotrobust}. We show that using the output of the DAE's hidden layer representation results in better computational efficiency and retains the effectiveness of the DAE output.  

Dimensionality reduction defense algorithms, on the other hand, include using Principal Component Analysis (PCA) \cite{pca}, traditional autoencoders \cite{ciss}, and a cascaded neural network architecture \cite{ciss} where the output of a DAE is input into a traditional autoencoder, whose bottleneck layer is extracted to produce a reduced dimensional representation of the input. PCA is computationally efficient but lacks robustness in withstanding the FGS attack since it does not account for the classification task, whereas the cascaded architecture proposed in \cite{ciss} serves as a robust defense but is computationally costly as it not only requires an additional autoencoder to be trained, but also reconstructs a compressed representation within its algorithm. Our proposed defense is shown to be as robust as the cascaded defense algorithm while reducing its accompanied computational cost. 

Another known defense strategy outside of the aforementioned categories of defense algorithms is adversarial retraining \cite{advtrn}, where the target classifier is retrained with simulated FGS attacks. Although adversarial retraining has worked as an effective defense in certain scenarios, it has been shown to be vulnerable to unforeseen black box attacks \cite{advtrnnotrobust}, even for very small perturbation magnitudes. However, for completeness, we compare our method to this widely proposed defense as well.

In this letter, we empirically show that using the output of the hidden layer of a Denoising AutoEncoder (DAE) is equivalently robust for defending the Fast Gradient Sign adversarial attack in comparison to the state-of-the-art, and requires much less computational power. We demonstrate the efficacy of our defense in both the semi-white box and black box scenarios. 

\vspace{-0.25cm}

\section{Methods and Setup}

\subsection{Datasets and Classifiers}
The empirical analysis in this letter is conducted using a supervised learning algorithm on two distinct ten-class datasets: the MNIST dataset of hand written digits \cite{mnist} and the Fashion-MNIST dataset of fashion articles \cite{fmnist}. Both datasets consist of  60,000 training samples and 10,000 disparate testing samples of 28$\times$28 gray scale images, denoted as $x \in {\rm I\!R}^n$. Each sample image is normalized to lie in $[0, 1]^n$, where $n = 28 \times 28 = 784$, and belongs to one of ten possible classes, $y \in \{0, 1, ...9\}$, which are one-hot encoded. In the MNIST dataset of hand written digits, each class corresponds directly to the same valued digit (i.e. the class `0' corresponds to the digit `0'). In the Fashion-MNIST dataset, each class corresponds to a different fashion article as follows: $\{0, 1, 2, 3, 4, 5, 6, 7, 8, 9\}$ = \{\emph{top, trouser, pullover, dress, coat, sandal, shirt, sneaker, bag, boot}\}. 

The target classifiers selected for classifying both of the foregoing datasets are identical in architecture. We use $F(\cdot): {\rm I\!R}^n \rightarrow \{0, 1\}^m$, to denote the classifier, i.e., $F(x)= \hat{y}$, where $\hat{y}$ is the predicted class from the classifier. $F(\cdot)$ is a fully connected artificial neural network with two hidden layers; consisting of 100 units each. The number of dimensions in the input is $n = $ 28 $\times$ 28 $=$ 784 and the number of dimensions in the output is $m=10$, which corresponds to one of ten possible classifications for each sample. Each unit in the hidden layers has a ReLU activation function and the output is a softmax layer. Each classifier is optimized using 100 epochs of the Adam algorithm \cite{adam} and a cross-entropy loss objective function. During training, each classifier uses a batch size of 200 and a learning rate of 0.001.  

\subsection{Attack Method}

In both the semi-white box and black box environments, the adversary generates adversarial perturbations, denoted as $\widetilde{x}$, according to the following formulation: 

\begin{equation}\label{fgs_eq} 
    \widetilde{x} = x + \eta * {sign}\left(\nabla_x J(w,x,y)\right).
\end{equation}

In (\ref{fgs_eq}), $x$ refers to the original input sample and ${sign}\left(\nabla_x J(w,x,y)\right)$ is the sign of the gradient of the cross-entropy cost function $J(w, x, y)$, which is a function of the input, $x$, the output classification, $y$, and the classifier weights, $w$. The parameter $\eta$ is the $l_\infty$-bounded perturbation, where $\eta \leq  \| \it{x} - \widetilde{x} \|_\infty$.

In the semi-white box threat model, the adversary has full knowledge of $F(\cdot)$ and is able to use its loss function to generate adversarial examples. However, in the black box environment, the adversary is forced to train an estimate model, denoted, $G(\cdot)$, which performs the same classification task as $F(\cdot)$. In this case, the adversary will attempt to mimic the behavior of the target classifier by constructing an artificial neural network architecture that is well fitted to the available training data. Therefore, $G(\cdot)$ is chosen to be a three-hidden-layer fully connected artificial neural network with 200, 200, and 100 units in each of its hidden layers, with respect to order from the input to output layers, followed by a ten-dimensional softmax output layer. We chose this architecture for the attacker's model as we found it to fit the training data well. Then, the adversary uses the loss function of $G(\cdot)$ to generate adversarial examples according to (\ref{fgs_eq}).

\subsection{Defense Method}

The defense consists of training a Denoising AutoEncoder (DAE) which consists of an encoder $p(x) = x'$: ${\rm I\!R}^n \rightarrow {\rm I\!R}^k$, which compresses the $n$-dimensional input into a $k$-dimensional representation ($k<n$), and a decoder $q(x') =\hat{x}$: ${\rm I\!R}^k \rightarrow {\rm I\!R}^n$, which reconstructs an approximation of the original $n$-dimensional input from the compressed $k$-dimensional representation. To train the DAE, the FGS attack is simulated on $F(x)$ using $\eta = 0.25$ on 30,000 randomly selected training samples and $\eta = 0.50$ on the remaining 30,000 training samples. The chosen bounds were empirically found to denoise the adversarial noise most effectively in the experimented perturbation range. The DAE is then trained to minimize the mean squared error between the clean input and reconstructed output:

\begin{equation}\label{dae} 
    \underset{p, q} {\text{Minimize }} \bigg|\bigg|\frac{1}{n} \sum_{i=1}^{n}(x_i - q(p({\tilde{x}_i}))\bigg|\bigg|^2.
\end{equation}

In addition to the 60,000 adversarial examples, the DAE is also trained to map the uncorrupted 60,000 training samples to their original representation to retain robustness in the absence of an attack. Finally, the encoder, $p(\cdot)$, is used to transform the training data and train a new classifier, $f(\cdot)$, of the same architecture as $F(\cdot)$ but with a $k$-dimensional input. Thus, the final defense for $F(\cdot)$ is $f(p(\cdot))$ 

Finding the optimal number of dimensions, $k$, to reduce the input data to is challenging and often dependent on the dataset. In this study, we explored compressing the input data to $k$ dimensions, where $k \in \{16, 32, 47, 64, 80, 94, 157\}$. For brevity, we have only included detailed results in this letter for $k=47$. However, our open source code\footnote{Code has been accepted for publication at: \url{https://codeocean.com/capsule/8113806/tree/v1}} contains results for all explored rates. 


\vspace{-0.25cm}

\section{Results}

We present the effectiveness of our defense at each perturbation magnitude by measuring the retained accuracy of the defense. The retained accuracy is the portion of samples that were correctly classified after the defense in relation to the number of samples that were initially classified correctly before the attack. Furthermore, we define the high SNR regime when $\eta < 0.25$ and the low SNR regime when $\eta > 0.25$.

\vspace{-0.25cm}

\subsection{Effectiveness of Defense: MNIST Digit Dataset}

\subsubsection{Semi-white Box Results}

In the semi-white box environment for the MNIST digit dataset, compressing the input sample to a 47-dimensional representation results in an average accuracy retention of 97.14\% across the noise range $\eta \in [0.01, 0.50]$. In comparison, the classifier is only able to retain an average of 15.95\% and 38.14\% of its accuracy across the same noise range when using no defense and adversarial retraining, respectively. Furthermore, at high SNR, using the hidden layer of the DAE provides an equivalently robust defense in comparison to the DAE output and the cascaded defense as shown in Fig. 1. However, the hidden layer defense is nearly 2.43x more computationally efficient in comparison to using the original $n$-dimensional representation as shown in Table 1. Finally, at low SNR, the hidden layer defense does result in a slight compromise in robustness, but this is accompanied by much less computational power.

\begin{figure}
\centerline{\includegraphics[width=\columnwidth]{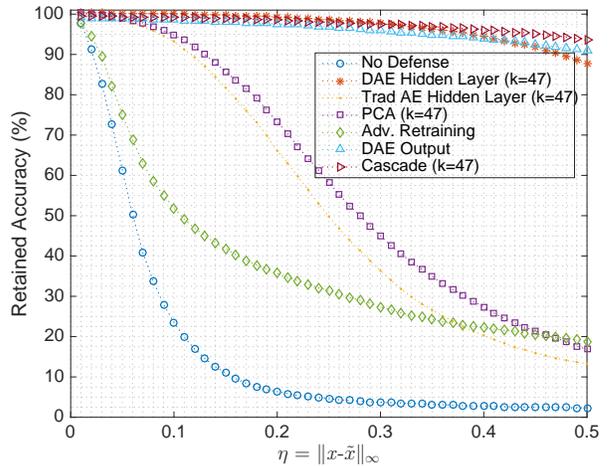}}
\caption{The retained accuracy of the MNIST digit dataset in the semi-white box environment for each tested perturbation magnitude.}
\end{figure}

\subsubsection{Black Box Results}


In the black box threat model, the DAE hidden layer defense achieves an average retention accuracy of 75.86\% in comparison to 19.16\% and 31.16\% when using no defense and adversarial retraining, respectively, for $\eta \in [0.01, 0.50]$. The robustness trend here, as shown in Fig. 2, is similar to that shown in the semi-white box model in that the DAE hidden layer defense is equivalently robust to the DAE output and the cascade. However, the lower accuracy retention in the low SNR regime for the black box scenario is more apparent in comparison the semi-white box scenario.

\begin{figure}
\centerline{\includegraphics[width=\columnwidth]{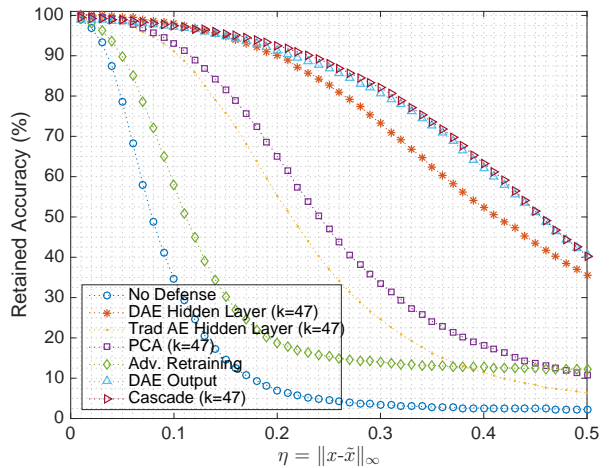}}
\caption{The accuracy of the MNIST digit dataset in the black box environment for each tested perturbation magnitude.}
\end{figure}

\subsection{Effectiveness of Defense: Fashion-MNIST Dataset}

\subsubsection{Semi-white Box Results}

As shown in Fig. 3, at high SNR, the DAE hidden layer defense is the most robust defense against the FGS attack for the Fashion-MNIST dataset in comparison to all other considered methods. As with the MNIST digit dataset, the Fashion-MNIST classification task experiences a slight compromise in accuracy in the low SNR regime which is accompanied by a significant reduction in computational cost. On average, the Fashion-MNIST classifier is able to retain an accuracy of 85.52\% for $\eta \in [0.01, 0.50]$ when using the DAE hidden layer defense, which slightly outperforms the DAE output defense that attains an average accuracy of 85.23\% in the same noise range, but delivers slightly lower robustness than the cascade, which attains an average accuracy of 89.34\% for $\eta \in [0.01, 0.50]$. Each of the foregoing defenses, however, significantly outperforms using no defense and adversarial retraining, which retain 17.33\% and 49.11\% accuracy on average, respectively.

\begin{figure}
\centerline{\includegraphics[width=\columnwidth]{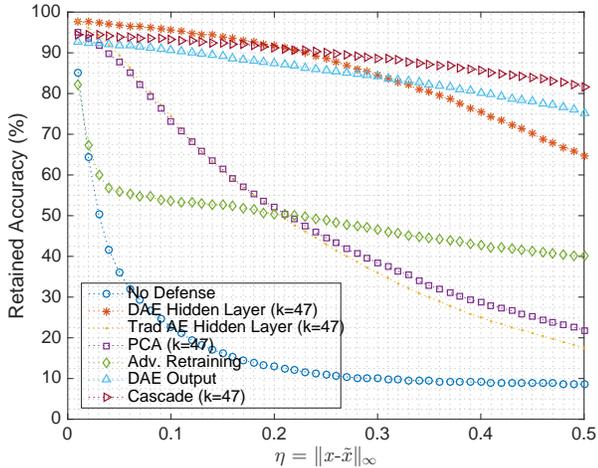}}
\caption{The accuracy of the Fashion-MNIST dataset in the semi-white box environment for each tested perturbation magnitude.}
\end{figure}

\subsubsection{Black Box Results}

In the black box threat model, the DAE hidden layer defense is able to attain an average retained accuracy of 63.30\%, which is significantly greater than the 26.47\% average retained accuracy of the mismatched attack without any defense. Similar to both threat models for the MNIST digit dataset, the DAE hidden layer defense outperforms, in terms of robustness, every other considered defense strategy in the high SNR regime, as shown in Fig. 4. However, the greater performance of the DAE hidden layer is more prominent in the high SNR regime of the Fashion-MNIST dataset in comparison to the high SNR regime of the MNIST digit dataset. Unlike the previous setups, adversarial retraining is able to retain an average accuracy of 63.06\% for $\eta \in [0.01, 0.50]$, which is approximately equivalent to using the DAE hidden layer defense. Although adversarial retraining outperforms the DAE hidden layer defense at low SNR, the latter defense is stronger in the high SNR regime where the attack is visually imperceptible and more important to defend.


\begin{figure}
\centerline{\includegraphics[width=\columnwidth]{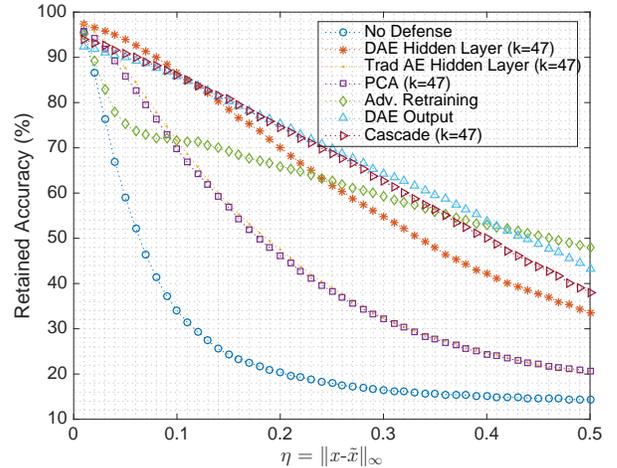}}
\caption{The accuracy of the Fashion-MNIST dataset in the black box environment for each tested perturbation magnitude.}
\end{figure}

\vspace{-0.15 cm}

\subsection{Effectiveness of Defense: Speedup}

Each experimented reduced input cardinality results in a significant speedup of over 2x in training the defense classifier, $f(\cdot)$, in comparison to training the target classifier, $F(\cdot)$, consisting of the initial 784-dimensional input. Table 1 shows the training time and speedup for various compression rates with the bold entries indicating the particular defense discussed throughout this letter for the MNIST digit dataset. The defense classifiers trained on the Fashion-MNIST dataset had almost identical speedup factors as those shown in Table 1, which was expected due to the classifiers' identical architectures. Although higher data compression results in greater speedup, it does not necessarily result in overall robustness (i.e., reducing input cardinality to $k=16$ results in a speedup of 2.67x and an average accuracy of 91.63\% whereas reducing to $k=157$ results in a greater average accuracy of 97.20\% but a slower speedup of 2.08x). Therefore, the optimal balance between compression and robustness must be found before implementing our proposed defense.


\vspace{-0.10 cm}
\section{Discussion and Future Work}


In the black box threat model, the gradient used to simulate adversarial perturbations differs from the gradient of the target classifier. As a result, the robustness of the defense is decreased due to the defense being trained to specifically filter adversarial noise proportional to the gradient of the target classifier. This trend is exemplified in Fig. 2, which in comparison to the results of Fig. 1, shows lower retained accuracy at low SNR. An adversary, as a result, can take advantage of this vulnerability and choose to generate an attack on a substitute model even if they have access to the target model. Upon discovering the adversary's new methodology, the defender would adapt their defense to variously crafted attacks, and this trend would continue alternating. In future work, we anticipate investigating, from a game theoretic perspective, how the knowledge held by the defender and attacker can give either party an advantage in defending and crafting an attack. 


\begin{table}
  \caption{Training Time and Speedups on MNIST Digit Dataset}
  \centering
  \begin{tabular}{|P{1.75cm}|P{1.75cm}|P{1.75cm}|P{1.75cm}|} 
    \hline
    Input Cardinality (k) & Training Time (s)          & Speedup (x) & Average Accuracy (\%)\\ \hline
    784                 & 120.67 & - & 96.16 \\ 
    157                      & 57.91 & 2.08x  & 97.20 \\ 
    94                   & 54.22 & 2.23x  & 96.83 \\ 
    80                   & 52.44 & 2.30x   & 96.86\\
    64                   & 52.26 & 2.31x   & 97.09\\
    \textbf{47}                   & \textbf{49.60} & \textbf{2.43x}  & \textbf{97.14} \\
    32                   & 46.61 & 2.59x  & 95.54 \\
    16                   & 45.27 & 2.67x   & 91.63\\
    \hline 

\end{tabular}
      \begin{tablenotes}
      \item The training time for each classifier along with the average retained accuracy for $\eta \in [0.01, 0.50]$ when using the DAE output defense ($k=784$), and when using various number of dimensions for the input through the DAE bottleneck layer as defense.
    \end{tablenotes}
\end{table}

\newpage

\bibliographystyle{IEEEtran}

\newpage
\section{Supplementary Material}
\subsection{Algorithms}
The defense used in our experiments consists of a fully connected multi-layer artificial neural network, which serves as the Denoising AutoEncoder (DAE). It is trained with five hidden layers containing 512, 256, $k$, 256, and 512 units, respectively. The number of dimensions at the most compressed hidden layer, denoted $k$, was used as an independent variable reflecting the various compression rates that were explored. Each DAE was trained using both clean and adversarial data, which was generated according to Algorithm 1. The DAE, which used no activation functions in its hidden layers and used a sigmoid activation function at the output layer, was optimized using the Adam algorithm with the mean squared error objective function using 150 epochs, a batch size of 256, and a learning rate of 0.001. During test time, all data was propagated through the defense shown in Algorithm 3. 

Every model was developed using the Keras Machine Learning package with a TensorFlow backend in Python, and the FGS attack was implemented using the Cleverhans Python package. We used an NVIDIA Tesla P100 GPU with 16 GB of memory for model training. 

\subsection{Datasets}

Fig. 5 and Fig. 6 show three samples from the MNIST digit dataset and the Fashion-MNIST Dataset, respectively, which are each initially classified correctly and then misclassified with high confidence upon the addition of an $l_{\infty}$ bounded error of $\eta = 0.10$.


\begin{algorithm}[tb]
   \caption{Attack Generator}
   \label{alg:attack}

    \bf{Function: }$attack$ \newline 
   {\bfseries Input:} $x$, $w$, $\eta$, $y$  \newline
   {\bfseries Result:} $\widetilde{x}$ \newline
  \hspace*{4mm} $\nabla_x J(w,x,y) = backpropagate(w, x, y)$\newline
  \hspace*{4mm} $\widetilde{x} = x + \eta * sign\left(\nabla_x J(w,x,y)\right)$\newline
  \hspace*{4mm} return $\widetilde{x}$

\end{algorithm}

\begin{algorithm}[tb]
   \caption{Defense Generator}
   \label{alg:defense}
 {\bfseries Input:} $x_{train}$, $y_{train}$, $w_{model}$ \newline
 {\bfseries Result:} $w_{DAE}, w_{defense}$ \newline
   
   \hspace*{4mm} $x^{(0.25)} = attack(x_{train}, w_{model}, 0.25, y_{train})$\hfill \newline
   \hspace*{4mm} $x^{(0.50)} = attack(x_{train}, w_{model}, 0.50, y_{train})$\newline
   \hspace*{4mm} $w_{DAE} = train\_DAE(x_{train}, x^{(0.25)}, x^{(0.50)})$\hfill \newline
   \hspace*{4mm} $x' = encoder_{w_{DAE}}(x_{train})$\hfill \newline
   \hspace*{4mm} $w_{defense} = train\_defense(x', y_{train})$\hfill \newline
   \hspace*{4mm}$return\; (w_{DAE}, w_{defense})$\newline
   
\end{algorithm}

\begin{algorithm}[tb]
   \caption{Real Time Defense}
   \label{alg:defense}
    
    {\bfseries Input:} ${x_{test}}$, $w_{DAE}$, $w_{defense}$\newline
   {\bfseries Result:} $\hat{y}$\newline
   \hspace*{4mm} $x_{test}' = encoder_{w_{DAE}}(x_{test})$\newline
   \hspace*{4mm} $\hat{y} = classify_{w_{defense}}(x_{test}')$ \newline 
   \hspace*{4mm}$return\; \hat{y}$\newline
   
\end{algorithm}

\setcounter{figure}{4}
\begin{figure}\label{fig:digits}
\centerline{\includegraphics[width=\columnwidth]{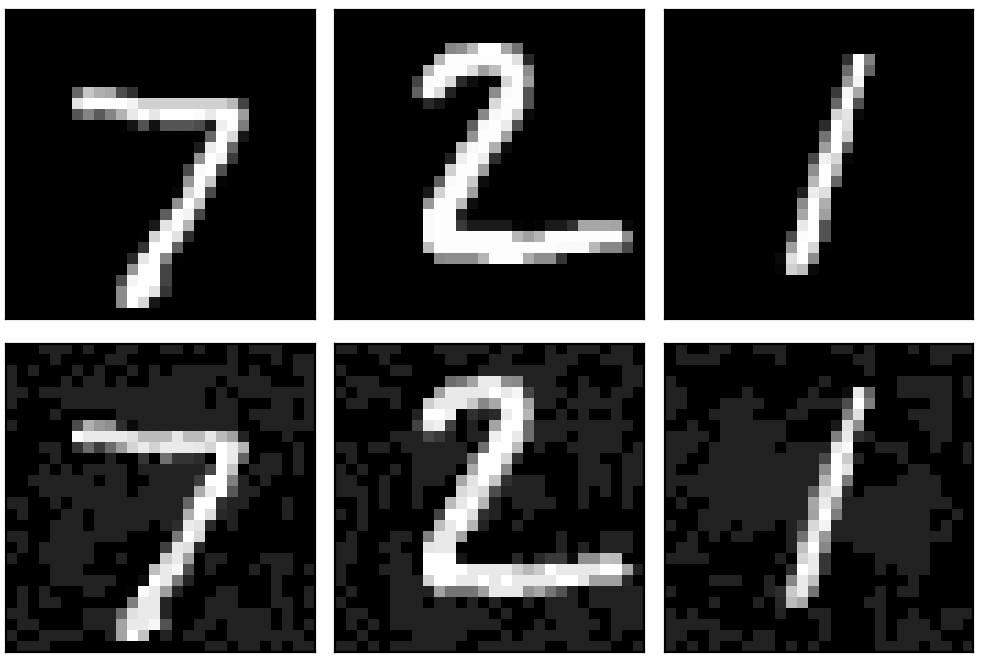}}
\caption{Example images from the MNIST digit dataset. Top row: original samples correctly classified as 7, 2, and 1. Bottom row: adversarially perturbed images now classified as 3, 1, and 8 with 100\%, 99\% and 100\% confidence, respectively.}
\end{figure}

\begin{figure}
\centerline{\includegraphics[width=\columnwidth]{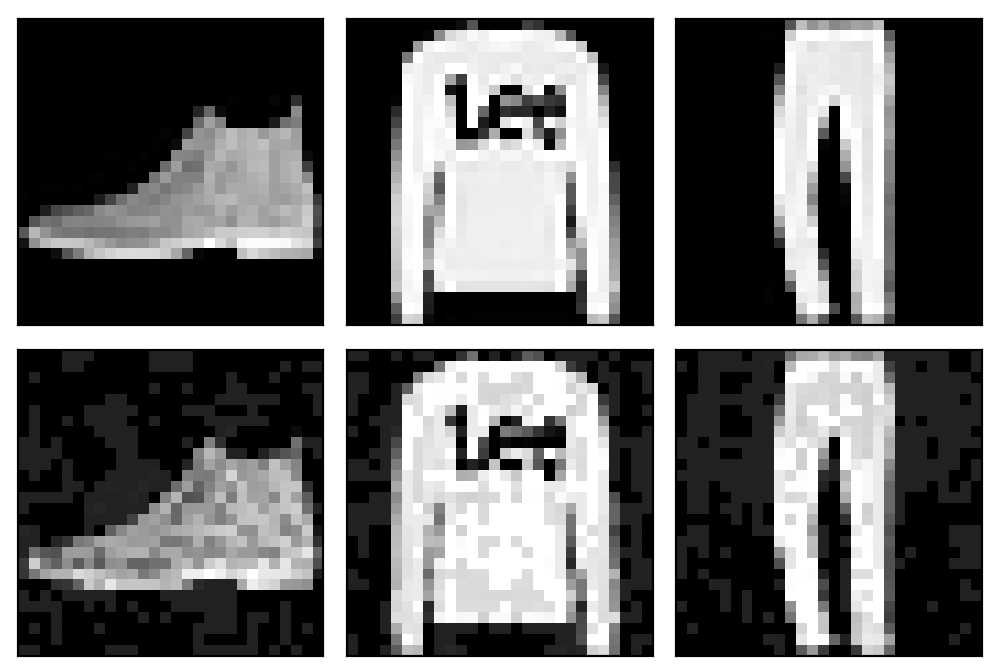}}
\caption{Example images from the Fashion-MNIST dataset. Top row: original samples correctly classified as ankle boot, pullover, and trouser. Bottom row: adversarially perturbed images now classified as sneaker, shirt, and coat; each with 100\% confidence.}
\end{figure}
\end{document}